\documentclass[sn-mathphys,Numbered]{sn-jnl}

\usepackage[utf8]{inputenc}
\usepackage[T1]{fontenc}    
\usepackage{subfigure}
\usepackage{amsmath,amssymb,amsfonts}

\DeclareMathOperator*{\argmin}{arg\,min}
\usepackage{xcolor}
\usepackage{url}
\usepackage{tabularx}

\usepackage{graphicx}%
\usepackage{multirow}%
\usepackage{amsthm}%
\usepackage{mathrsfs}%
\usepackage[title]{appendix}%
\usepackage{textcomp}%
\usepackage{manyfoot}%
\usepackage{booktabs}%
\usepackage{algorithm}%
\usepackage{algorithmicx}%
\usepackage{algpseudocode}%
\usepackage{listings}%

\theoremstyle{thmstyleone}%
\theoremstyle{thmstyletwo}%

\theoremstyle{thmstylethree}%

\begin{document}

\title{Lyapunov-Guided Representation of Recurrent Neural Network Performance}


\author[1]{\fnm{Ryan} \sur{Vogt}}\email{ravogt95\text{@}uw.edu}

\author[2]{\fnm{Yang} \sur{Zheng}}\email{zheng94\text{@}uw.edu}

\author*[1,2]{\fnm{Eli} \sur{Shlizerman}}\email{shlizee\text{@}uw.edu}

\affil*[1]{\orgdiv{Department of Applied Mathematics}, \orgname{University of Washington}, \orgaddress{\city{Seattle}, \postcode{98195}, \state{WA}, \country{US}}}

\affil*[2]{\orgdiv{Department of Electrical and Computer Engineering}, \orgname{University of Washington}, \orgaddress{\city{Seattle}, \postcode{98195}, \state{WA}, \country{US}}}


\abstract{
Recurrent Neural Networks (RNN) are ubiquitous computing systems for sequences and multivariate time series data. While several robust architectures of RNN are known, it is unclear how to relate RNN initialization, architecture, and other hyperparameters with accuracy for a given task. In this work, we propose to treat RNN as dynamical systems and to correlate hyperparameters with accuracy through Lyapunov spectral analysis, a methodology specifically designed for nonlinear dynamical systems. To address the fact that RNN features go beyond the existing Lyapunov spectral analysis, we propose to infer relevant features from the Lyapunov spectrum with an Autoencoder and an embedding of its latent representation (AeLLE). Our studies of various RNN architectures show that AeLLE successfully correlates RNN Lyapunov spectrum with accuracy. Furthermore, the latent representation learned by AeLLE is generalizable to novel inputs from the same task and is formed early in the process of RNN training. The latter property allows for the prediction of the accuracy to which RNN would converge when training is complete. We conclude that representation of RNN through Lyapunov spectrum along with AeLLE provides a novel method for organization and interpretation of variants of RNN architectures.

}

\maketitle


\section*{Introduction}

Recurrent Neural Networks (RNN) specialize in processing sequential data, such as natural speech or text, and broadly, multivariate time series data~\cite{Pang_2019_CVPR, mikolov2010recurrent, noisy1998}. With such omnipresent data, RNN address a wide range of tasks in the form of prediction, generation, and translation for applications which include stock prices, markers of human body joints, music composition, spoken language, and sign language ~\cite{tino2001financial, su2020predict, Pennington2017, sutskever2014sequence, gregor2015draw, choi2016text, mao2018deepj, guo2018hierarchical}. While RNN are ubiquitous systems, these networks cannot be easily explained in terms of the architectures they assume, the parameters that they incorporate, and the learning process that they undergo. As a result, it is not straightforward to associate an optimally performing RNN with a particular dataset and a particular task. The difficulty stems from RNN characteristics making them intricate dynamic systems. In particular, RNN can be classified as \textit{(i) nonlinear} \textit{(ii) high-dimensional} \textit{(iii) non-autonomous}  dynamical systems with \textit{(iv) varying parameters}, which are either global (hyperparameters) or trainable weights (connectivity). 

Architectural variants of RNN that are more optimal and robust than unstructured RNN have been introduced. From the long-established gated RNN, such as Long Short-Term Memory (LSTM)~\cite{hochreiter1997long} and Gated Recurrent Units (GRU)~\cite{cho2014gru, chung2014empiricalgru}, to more recent networks, such as Anti-Symmetric RNN (ASRNN)~\cite{Chang2019}, Orthogonal RNN (ORNN) ~\cite{vorontsov2017orthogonality}, Coupled Oscillatory RNN (coRNN)~\cite{rusch2021coupled}, Lipschitz RNN (LRNN)~\cite{erichson2020lipschitz} and more. While such specific architectures are robust, they represent singular points in the space of RNN. In addition, the accuracy of these architecture still depends on hyperparameters, among which are the initial state of the network, initialization of the network connectivity weights, optimization settings, architectural parameters, and the input statistics. Each of these factors has the potential to impact network learning and as a consequence task performance. Indeed, by fixing the task and randomly sampling hyperparameters and inputs, the accuracy of otherwise identical networks can vary significantly, leading to potentially unpredictable behavior.

A powerful dynamical system method for characterization and predictability of dynamical systems is \textit{Lyapunov Exponents (LE)}~\cite{ruelle1979ergodic, Oseledets:2008}. Recent work identifying RNN as dynamical systems has extended LE calculation and analysis to these systems~\cite{Engelken}, but the connection between LE and network performance has not been explored extensively.
Examples in~\cite{vogt2020lyapunov} suggest that features of LE spectrum are correlated with RNN robustness and accuracy, but standard features such as maximum and mean LE can have weak correlation, making consistent characterization of network quality using a fixed set of LE features unfeasible.

Such inconsistency motivates this work where we develop a data driven methodology, called \textit{AeLLE}, to infer LE spectrum features and associate them with RNN performance. The methodology implements an Autoencoder (Ae) which learns through its \textit{L}atent units a representation of \textit{LE} spectrum (LLE) and correlates the spectrum with the accuracy of RNN for a particular task. The latent representation appears to be low dimensional such that even a simple linear embedding of the representation, denoted as AeLLE, corresponds to a classifier for the selection of optimally performing parameters of RNN based on LE spectrum. We show that once AeLLE is trained, it holds for novel inputs and we also investigate the correlation between AeLLE classification accuracy and the need for RNN training.

The significance of the proposed AeLLE method is that it is a novel LE embedding that effectively facilitates interpretation and classification for a variety of RNN models. For example, we show that AeLLE separates high- and low-accuracy networks across varying weight initialization hyperparameter, network size, and network architecture with no knowledge about the underlying network besides its Lyapunov spectrum. Notably, such separation is not observed when AeLLE is omitted and Lyapunov spectrum is directly used in conjunction with standard embeddings. Indeed, our results indicate that AeLLE representation is a necessary step which follows LE spectrum. While it requires additional computational power and time due to the training of the autoencoder, it is a necessary computational step since it is able to identify the subtle features of the Lyapunov spectrum. As a result, AeLLE implicitly identifies the network dynamics, which are integral to performance across a wide variety of network hyperparameters. Furthermore, we show that AeLLE can be used to predict final network accuracy early in training. This suggests that during training AeLLE recovers the Lyapunov spectrum features which are instrumental for trainability. These features are not evident in standard LE statistics, as discussed in~\cite{vogt2020lyapunov,mikhaeil2022difficulty,herrmann2022chaotic}.

\section*{Related Work}

\subsection*{Spectral Analysis and Model Quality}
Spectral methods have been used to characterize information propagation properties, and thus model quality, of RNNs. Since vanishing and exploding gradients arise from long products of Jacobians of the hidden states dynamics whose norm could exponentially grow or decay, much effort has been made to mathematically describe the link between model parameters and the eigen- and singular-value spectra of long products~\cite{Pascanu,Poole2016, wang2018predicting,Chen2018}. For architectures used in practice, these approaches appear to have a limited scope~\cite{yang2019scaling,zheng2020r}. This is likely due to spectra having non-trivial properties reflecting intricate long-time dependencies within the trajectory, and due to the need to take into account the dynamic nature of RNN systems, which is the reason for proposing to use Lyapunov Exponents in this work for such characterization.

Other techniques for inferring model quality include performing spectral analysis of deep neural network weights and fitting their distribution to truncated power laws to correlate with performance \cite{martin_trends}. Another approach using Koopman operators to linearize RNNs demonstrates that linear representations of RNNs capture the dominant modes of the networks and are able to achieve comparable performance to their non-linear counterparts \cite{naiman_koopman}.

\subsection*{Dynamical Systems Approaches to RNNs}
The identification of RNN as dynamical systems and as such developing appropriate analyses appears as a prospective direction. Recently, dynamical systems methodology has been applied to introduce constraints to RNN architecture to achieve better robustness, such as orthogonal (unitary) RNN~\cite{wisdom2016full,jing2017tunable,mhammedi2017efficient,vorontsov2017orthogonality,azencot2021differential,Chang2019,kerg2019non} and additional architectures such as coupled oscillatory RNN~\cite{rusch2021coupled} and Lipschitz RNN~\cite{erichson2020lipschitz}. These approaches set network weights to form dynamical systems which have the desired Jacobians for long-term information propagation. In addition, analyses such as stochastic treatment of the training procedure have been shown to stabilize various RNN~\cite{lim2021noisy}. Furthermore, a universality analysis of fixed points of different RNN, proposed in~\cite{maheswaranathan2019universality}, hints that RNN architectures could be organized in similarity classes such that despite having different architectural properties would exhibit similar dynamics when optimally trained. In light of this analysis, it remains unclear to which similarity classes in the space of RNN the constrained architectures belong and what is the distribution of the architectures within each class. These unknowns warrant development of dynamical systems tools that \textit{characterize and classify RNN variants}.

\subsection*{Lyapunov Spectrum}
\textit{Lyapunov Exponents}~\cite{ruelle1979ergodic, Oseledets:2008} capture the information generation by a system's dynamics through  measurement of the separation rate of infinitesimally close trajectories. The number of Lyapunov exponents of a system is equal to the dimension of that system, and the collection of all LE is called LE spectrum. The maximal LE will determine the linear stability of the system~\cite{perron1930ordnungszahlen}. Further, a system having LE greater than zero represents chaotic dynamics with the magnitude of the first exponent indicating the degree of chaos; when the magnitude decreases and approaches zero, the degree of chaos decreases as well. When all exponents are negative, the dynamics converge towards a fixed point attractor. Zero exponents represent limit cycles or quasiperiodic orbits~\cite{PhysRevLett.73.1927, Abarbanel1991}. Additional features of LE, even non-direct, correspond to properties of the dynamical system. For example, the mean exponent determines the rate of contraction of full volume elements and is similar to the KS entropy~\cite{SHIBATA2001182}. The LE variance measures heterogeneity in stability across different directions and can reflect the conditioning of the product of many Jacobians.

When LE are computed from observed evolution of the system, Oseledets theorem guarantees that LE characterize each ergodic component of the system, i.e., when long enough evolution of a trajectory in an ergodic component is sampled, the computed LE spectrum is guaranteed to be the same (see Methods section for details of computation)~\cite{ergodic-components, ochs1999stability}.  Efficient approaches and algorithms have been developed for computing LE spectrum~\cite{lyap-algos}. These have been applied to various dynamical systems including hidden states of RNN and variants such as LSTM and GRU~\cite{Engelken}. This approach relies on the theory of random dynamical systems which establishes LE spectrum even for a system driven by a noisy random input sequence sampled from a stationary distribution~\cite{arnold1995random}. It has been demonstrated that some features of LE spectra can have meaningful correlation with the performance of the corresponding networks on a given task, but the selection of relevant features of it is task-dependent and challenging to determine a priori~\cite{vogt2020lyapunov}.

\section*{Methods}
The proposed AeLLE methodology consists of three steps: 1) Computation of LE spectrum, 2) Autoencoder for LE spectrum, and 3) Embedding of Autoencoder Latent representation.

\subsection*{Computation of LE~\cite{Engelken,vogt2020lyapunov}}
We compute LE by adopting the well-established  algorithm~\cite{Benettin1980,Dieci:1995gp} and follow the implementation in~\cite{Engelken,vogt2020lyapunov}. 
For a particular task, each batch of input sequences is sampled from a set of fixed-length sequences of the same distribution. We choose this set to be the validation set. For each input sequence in a batch, a matrix $\textbf{Q}_0$ is initialized as the identity to represent an orthogonal set of nearby initial states whose evolution will be tracked in the sequence of matrices $\textbf{Q}_t$. The hidden states \textbf{$h_t$} are initialized as zeros. 

To track the expansion and the contraction of the vectors of $\textbf{Q}_t$, the Jacobian of the hidden states at step t, $\textbf{J}_{t}$, is calculated and then applied to the vectors of $\textbf{Q}_t$. The Jacobian $\textbf{J}_t$ can be found by taking the partial derivatives of the RNN hidden states at time $t$, $h_t$, with respect to the hidden states at time at $t-1$, $h_{t-1}$
\begin{equation}
    \left[\textbf{J}_{t}\right]_{ij} = \frac{\partial \textbf{h}_{t}^j}{\partial \textbf{h}_{t-1}^i}.
\end{equation}

Beyond the hidden states, the Jacobian will depend on the input $x_t$ to the network. This dependence allows us to capture dynamics of a network as it responds to input. The expansion factor of each vector is calculated by finding the corresponding R-matrix that results from updating \textbf{Q} when computing the \textit{QR} decomposition at each time step
\begin{equation}
    \textbf{Q}_{t+1}, \textbf{R}_{t+1} = QR(\textbf{J}_{t}\textbf{Q}_{t}).
\end{equation}
If $r_t^i$ is the expansion factor of the $i^{th}$ vector at time step $t$ -- corresponding to the $i^{th}$ diagonal element of \textbf{R} in the QR decomposition-- then the $i^{th}$ LE $\lambda_i$ resulting from an input signal of length $T$ is given by
\begin{equation}\label{eq:Lyap_sum}
    \lambda_k = \frac{1}{T}\sum_{t=1}^T {\log}(r_t^k)
\end{equation}
The LE resulting from each input $x^m$ in the batch of input sequences are calculated in parallel and then averaged. For each experiment, the LE were calculated over a fixed number of time steps with \textit{n} different input sequences. The mean of \textit{n} resulting LE spectra is reported as the LE spectrum. To normalize the spectra across different network sizes and consequently the number of LE in the spectrum, we interpolate the spectrum such that it retains the shape of the largest network size. Through this interpolation, we can represent the LE spectra as curves. Spectra curves will have the same number of LE points for small networks and for larger networks.

The expressions for the Jacobians $\mathbf{J}_t$ used in these calculations can be found in the Supplemental Materials.

\subsection*{Autoencoder for LE spectrum}
An autoencoder consists of two components: an \textit{encoder} network $\mathcal{\phi}$ which transforms the input into a representation in the latent layer, and a \textit{decoder} network $\mathcal{\psi}$ which transforms the latent representation into a reconstruction of the original input. 
Over the course of training, the Latent layer becomes representative of the variance in the input data and extracts key features that might not immediately be apparent in the input. 

In addition to the reconstruction task, it is possible to include constraints on the optimization by formulating of a loss function for the Latent layer values (Latent space), e.g., a classification or prediction criterion. This can constrain the organization of values in the Latent space~\cite{goodfellow2016deep, chollet2021deep}.
\begin{figure*}[t!]
    \centering
    \includegraphics[width = 0.85\textwidth]{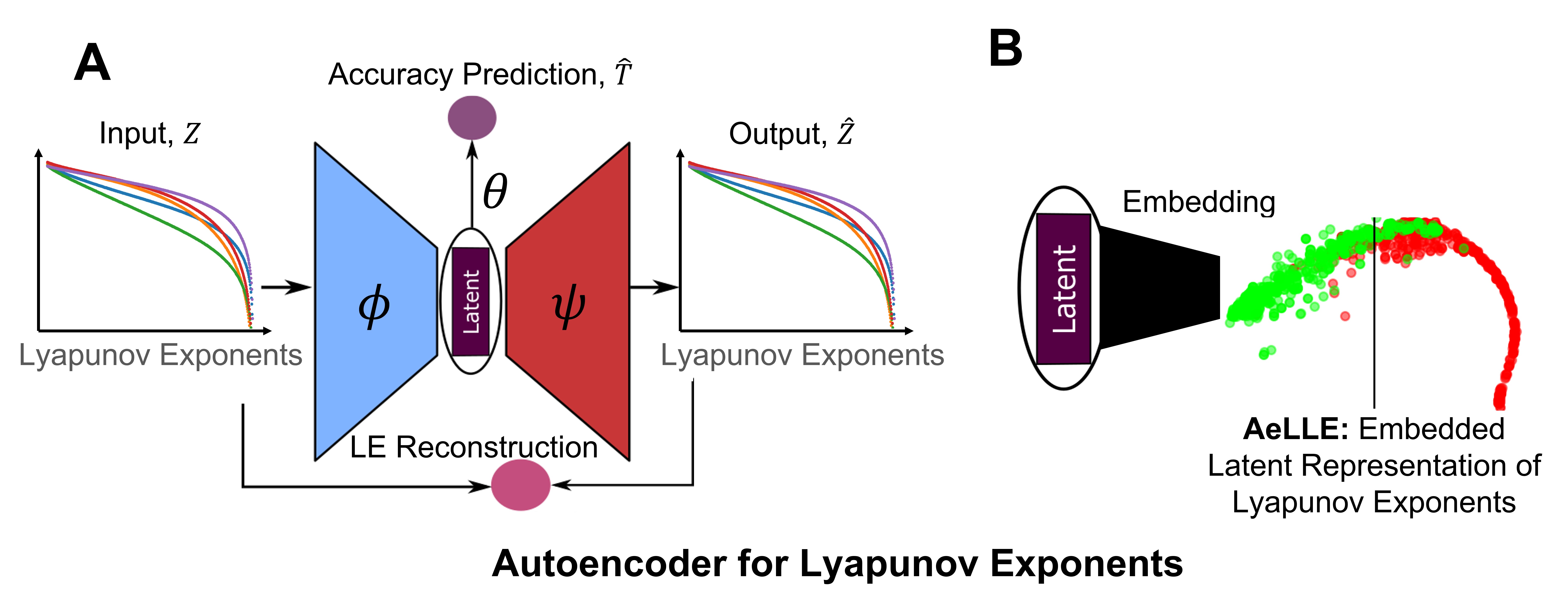}
    \caption{\textbf{AeLLE: LE spectrum Autoencoder and Latent Representation Embedding.} A) The autoencoder takes Lyapunov Exponents as input. This input is then embedded into a latent space (purple) by the encoder (blue). From this latent space, the autoencoder predicts the accuracy of the corresponding network, which is compared to the true accuracy of the network to get the prediction loss. Simultaneously, the Lyapunov spectrum is reconstructed from this latent space by the decoder(red). These reconstructed Lyapunov Exponents are then compared to the input Lyapunov exponents to get the reconstruction loss. These sums are added together with normalization factor $\alpha$ to get the total loss.  B) The Latent space of the LE spectrum Autoencoder correlates LE spectrum and accuracy. Embedding of Latent space representation provides a low-dimensional clustering and classification space, leading to separation between high-accuracy (green) and low-accuracy (red) networks. We find that Latent space clusters the LE spectra well such that simple embedding (PCA) and simple linear classifiers (hyperplane, hyperellipse or threshold) classify RNN variants according to accuracy. Shown is an example linear classifier along the first PC dimension for networks trained on the CharRNN task (see Results section for more details).}
    \label{fig:Methods}
\end{figure*}
We propose an adapted Autoencoder methodology for correlating LE spectra and RNN task accuracy. In this setup, we consider the LE spectrum as the input \emph{\textbf{Z}}. Our autoencoder consists of a fully-connected encoder network $\mathcal{\phi} $, a fully-connected decoder network $\mathcal{\psi}$, and an additional linear prediction network $\xi$ defined by
\begin{equation}\label{eq:AE}
    \begin{split}
        \hat{Z} = (\psi \circ \phi)Z, \\
        \hat{T} = (\xi \circ \phi)Z,
    \end{split}
\end{equation}
where \emph{$\hat{\textbf{Z}}$}, and \emph{$\hat{\textbf{T}}$}, correspond to the output from the decoder and predicted accuracy, respectively, with loss of $L =  \|Z - \hat{Z} \|^2 + \alpha \cdot \| T - \hat{T}\|_l$. Ae performs the reconstruction task, optimization of the first term of Eq.~\ref{eq:AEloss}, mean-squared reconstruction error of LE spectrum, as well as prediction of the associated RNN accuracy \emph{\textbf{T}} (best validation loss), the second term of Eq.~\ref{eq:AEloss}.
\begin{equation}\label{eq:AEloss}
    \phi, \psi, \xi =  \argmin_{\phi, \psi, \xi} (\|Z - \hat{Z} \|^2 + \alpha \cdot \| T - \hat{T}\|_l ).
\end{equation}
The parameter $l$ can defined based on the desired behavior. The most common choices are $l=1$, indicating the 1-norm, and $l=2$, indicating the 2-norm.

During training of Ae, the weight $\alpha$ in the prediction loss is gradually being increased so that Ae emphasizes RNN error prediction once the reconstruction error has converged. We found that this approach allows to capture features of both RNN dynamics and accuracy. A choice of $\alpha$ being too small leads to dominance of the reconstruction loss such that the correlation between LE spectrum and RNN accuracy is not captured. Conversely, when $\alpha$ is initially set to a large value, the reconstruction along with the prediction diverge. The convergence of Ae for different RNN variants, as we demonstrate in Results section, shows that correlative features between LE spectrum and RNN accuracy can be inferred. 
The dependency of Ae convergence on a delicate balance of the two losses reconfirms that these features are tangled and thus the need for Ae embedding. We describe the settings of $\alpha$ and additional Ae implementation details in Supplementary Materials.

\subsection*{Embedding of Autoencoder Latent Representation}
When the loss function of Ae converges, it indicates that the Latent space captures the correlation between LE spectrum and RNN accuracy. However, an additional step is typically required to achieve an organization of the Latent representation based on RNN variants accuracy. For this purpose, a low dimensional embedding, denoted as AeLLE, of the Latent representation needs to be implemented. An effective embedding would indicate the number of dominant features needed for the organization, provide a classification space for the LE spectrum features, and connect them with RNN parameters. We propose to apply the Principal Component Analysis (PCA) embedding to the Latent representation~\cite{pearson1901liii, su2020clustering}. The embedding consists of performing Principal Component Analysis and projecting the representation on the first few Principal Component directions (e.g. 2 or 3). While other, nonlinear, embeddings are possible, e.g., tSNE or UMAP~\cite{van2008visualizing, mcinnes2018umap}, the simple linear projection onto the first two principal components of the latent space results in an effective organization. This indicates that the Latent representation has successfully captured the characterizing features of performance. 
We show in the Results section that, for all examples of RNN architectures and tasks that we considered, the PCA embedding is sufficient to provide an effective space. In particular, in this space, most accurate RNN variants (green) can be separated from other variants (red) through a simple clustering procedure. 

\section*{Results}

To investigate the applications and generality of our proposed method, we consider tasks with various inputs and outputs and various RNN architectures that have been demonstrated as effective models for these tasks. In particular, we choose three tasks: Signal Reconstruction, Character Prediction and Sequential MNIST. All three tasks involve learning temporal relations in the data with different forms of the input and objectives of the task. Specifically, the inputs range from low-dimensional signals to categorical character streams to pixel greyscale values. Nonetheless, across this wide variety of inputs and tasks, AeLLE space and clustering is consistently able to separate variants of hyperparameters according to accuracy in a way that is more informative than network hyperparameters alone. 

More specifically, we consider \textit{(i)} \textit{Signal Reconstruction} task, also known as target learning. In this task, a random RNN is being tracked to generate a target output signal from a random input~\cite{sussillo2009generating, zheng2020r}. This task involves intricate time-dependent signals and a generic RNN for which dynamics in the absence of training are chaotic. With this example, we demonstrate that our method is able to distinguish between networks of high and low accuracy across \textit{initialization parameters}.

\textit{(ii) Character Prediction} is a common task which takes as an input a sequence of textual characters and outputs the next character of the sequence. This task is a rather simple task and is used to benchmark various RNN variants. With this task, we demonstrate that our method is able to distinguish across \textit{network sizes}, in addition to initialization parameters.

\textit{(iii) Sequential MNIST} is a more extensive benchmark for RNN classification accuracy. The input in the task is an image of a handwritten digit unrolled into a sequence of numerical values (pixels' greyscale values) and the output is a corresponding label of the digit. We investigate accuracy of various RNN variants on row-wise SMNIST, demonstrating that our method distinguishes according to performance across \textit{network architectures}. We describe the outcomes of AeLLE application and resulting insights per each task below.

For each of these tasks, we present the low-dimensional projection of AeLLE using the first two principal components. Furthermore, we reduce the projection to a single dimension by showing the distribution of the AeLLE in the first principal component as stacked histograms of the high- and low-accuracy networks or the different hyperparameters. For these stacked histograms, we represent each distribution separately and stack the histograms on top of each other, such that the bar heights are not cumulative and are discrete. Therefore, the ordering of the colors will always be the same, and the height of each bar indicates the number of networks in that bin, regardless of whether it is above or below the other colors.


\begin{figure*} [t!]
    \centering
    \includegraphics[width=\textwidth]{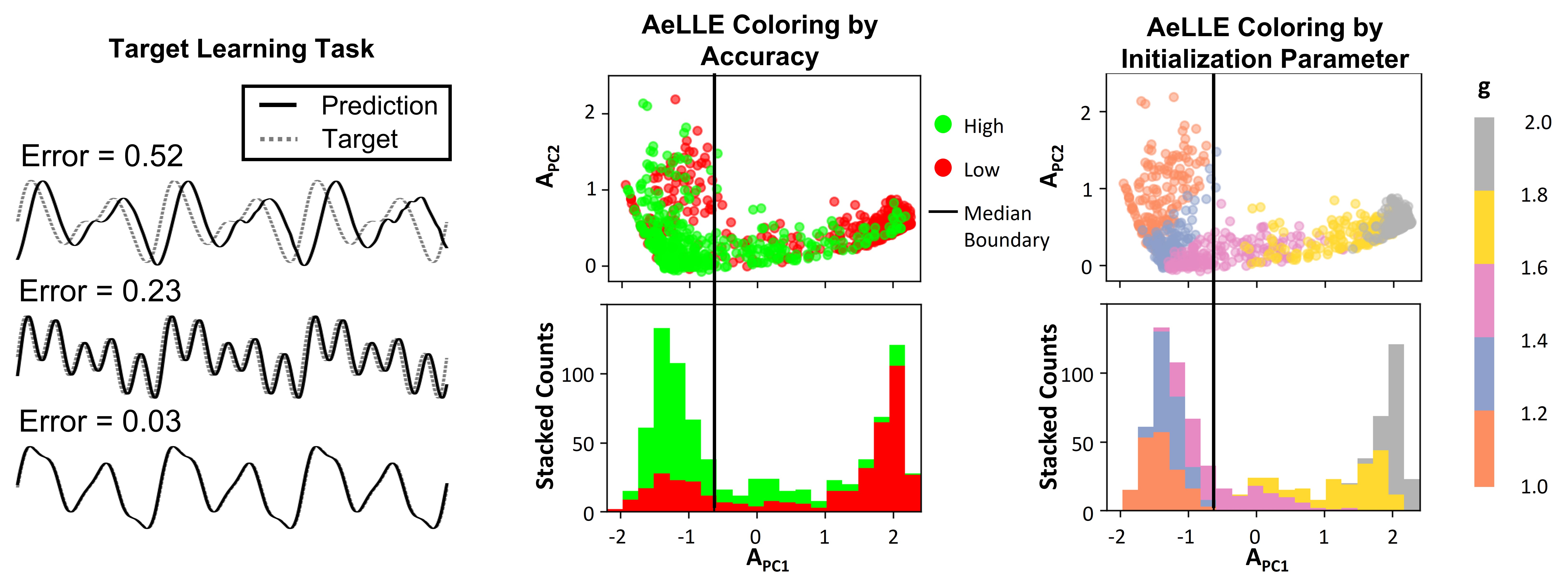}
        \caption{\textbf{Clustering by performance across initialization parameter: AeLLE for networks trained on Signal Reconstruction Task.} \textit{Left:} The signal reconstruction task involves recreating a target signal. Higher losses indicate a greater pointwise difference between the target and predicted signals. \textit{Center:} In the AeLLE representation, most of the high-accuracy networks (green) cluster to the left of most of the low-accuracy networks (red). The black vertical line indicates the location of the median classifier along the first principal component of the AeLLE space. Moreover, the greatest concentration of the high-accuracy networks is in the bottom-left of the space shown, which is consistent with the stacked histogram indicating that there are relatively few high-accuracy networks to the right of the median compared to low-accuracy, and many more to the left. \textit{Right:} In comparison, the cluster to the bottom-left of the space shown contains a mixture of networks with initialization parameters ranging from 1.0 to 1.6. In general, networks with larger initialization parameter $g$ (particularly once $g>1.6$), tend to cluster further to the right in this space. This is consistent with the fact that networks with $g>1.6$ tend to have lower accuracy on this task, but networks with $1<g<1.6$ tend to have similarly high accuracy (see Supplemental Materials).}
    \label{fig:Target_Learning}
\end{figure*}

\subsection*{Signal Reconstruction via Target Learning with Random RNN}

To examine how AeLLE interprets generic RNN with time evolving signals as output and input, we test Rank-1 RNN. Such a model corresponds to training a single rank of the connectivity matrix, the output weights $W$, on the task of target learning
. We set the target signal (output) to be a four-sine wave, a benchmark used in~\cite{sussillo2009generating}. A key parameter in Rank-1 RNN is the amplification factor of the connectivity $g$ which controls the output signal in the absence of training. For $g \leq 1$, the output signal is zero, while for $g \geq 1.8$ the output signal is strongly chaotic. In the interval $1< g < 1.8$ the output signal is weakly chaotic. Previous work has shown that the network can generate the target signal when it is in the weakly chaotic regime, i.e., $1<g<1.8$ and trained with FORCE optimization algorithm. Moreover, it has been shown that this training setup most consistently optimally converges when the amplification factor \textit{g} is in the interval $1.2<g<1.5$ ~\cite{sussillo2009generating, depasquale2018full}.

However, not all samples of the random connectivity correspond to accurate target generation. Even for $g$ values in the weakly chaotic interval, there would be Rank-1 RNN variants that fail to follow the target. Thereby, the target learning task, Rank-1 RNN architecture, and FORCE optimization are ideal candidates to test whether AeLLE can organize the variants of Rank-1 RNN models according to accuracy. 

The candidate hyperparameters for variation would be of 1) samples of fixed connectivity weights (from normal distribution) and 2) the parameter $g$ within the weakly chaotic regime. We structure the benchmark set to include 1200 hyperparameter variants and compute LE spectrum for each of them. After training is complete, LE in the validation set are projected onto the Autoencoder's AeLLE space, depicted in Fig.~\ref{fig:Target_Learning}, where each sample is a dot in the $A_{PC_1}-A_{PC_2}$ plane. 


Our results show that AeLLE organizes the variants in a 2D space of $A_{PC_1}-A_{PC_2}$ according to accuracy. The variants with smaller error values (high accuracy) ($<0.57$) are colored in \textit{green} and variants with larger error values (low accuracy) ($>0.57$) are colored in \textit{red}. We demonstrate the disparity in the signals that different error values correspond to in Fig.~\ref{fig:Target_Learning}-left. AeLLE space succeeds to correlate LE spectrum with accuracy such that most low-error networks are clustered in the bottom-left of the two-dimensional projection (see Fig. \ref{fig:Target_Learning}-center), whereas large-error networks are concentrated in primarily to the right and top of the region shown. This clustering of high-accuracy networks allows for the identification of multiple candidates as top performing variants in this space. Comparison of AeLLE clustering with a direct clustering according to values of $g$, Fig.~\ref{fig:Target_Learning}-right, shows that while most networks with $g< 1.7$ include variants with low-error, there are also variants with high-error for each value of $g$. This is not the case for all variants in the low-error hyperellipse of the AeLLE space. These variants have different $g$ and connectivity values and sampling from the hyperellipse provides a higher probability Rank-1 RNN variant to be accurate.

\begin{figure*} [t!]
    \centering
    \includegraphics[width=\textwidth]{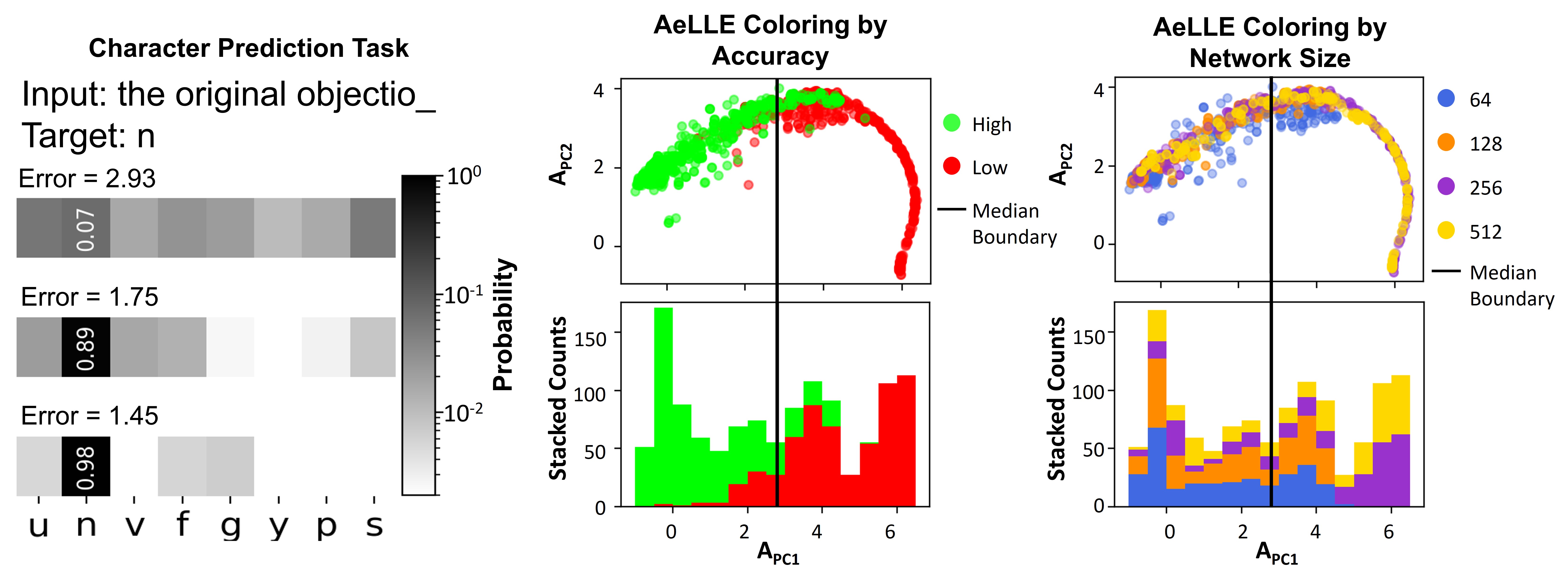}
        \caption{\textbf{Clustering by performance across network size: AeLLE for LSTM RNNs trained on CharRNN task.} \textit{Left:} The CharRNN task involves predicting the next character in a sequence. Larger losses correspond to less confidence in predicting the next character in a sequence. \textit{Center:} In the AeLLE representation the higher-accuracy networks, regardless of size, tend to cluster by performance, with the low-accuracy networks to the left in this representation. \textit{Right:} By contrast, LSTM of different sizes are often mixed together in this representation, with larger networks covering a wider range within the AeLLE space, but still overlapping with smaller networks throughout the space.}
    \label{fig:CharRNN}
\end{figure*}

Notably, AeLLE contributes to this problem beyond validation of AeLLE itself. The representation selects variants that are “non-trivial” as well, i.e., it is interesting to examine the particular configurations of models that do not belong to $1.2<g<1.5$  but AeLLE still rightfully reports them being successful at target learning or vice versa models in $1.2<g<1.5$ but do not reconstruct the target. AeLLE indeed identifies such models that are on the tail of the distribution in terms of the parameter \textit{g}.  

For comparison, we calculated the F1 score of the classifier which uses the median first principal component value as the decision boundary for classifiers which use simple statistics of the Lyapunov spectra. When we use the median value of the Lyapunov spectrum means and maximum Lyapunov exponent as the decision boundaries, the resulting F1 scores are 0.705 and 0.504, respectively, meaning that the mean LE is much more indicative of performance than the maximum LE for this task. Additionally, we define another classifier by projecting the raw Lyapunov exponents onto their first two principal components and using the median of the first principal component as the decision boundary to get a LE PC classifier. For this task, we find the LE PC classifier achieves similar performance to the LE mean, with an F1 score of 0.703.
Meanwhile, the AeLLE classifier achieves an F1 score of 0.724, indicating a significant improvement over the max LE classifier and a modest improvement over the mean LE and LE PC classifiers (see Supplementary Materials for more details).


\subsection*{Character Prediction with LSTM RNNs of Different Size}

Multiple RNN tasks are concerned with non-time dependent signals, such as sequences of characters in a written text.
Therefore, we test AeLLE on LSTM networks that perform the character prediction task (CharRNN) in which for a given sequence of characters (input) the network predicts the character (output) that follows. In particular, we train LSTM networks on English translation of Leo Tolstoy's \textit{War and Peace}, similar to the setup described in \cite{karpathy2015visualizing}. In this setup, each unique character is assigned an index (number of unique characters in this text is 82), and the text is split into disjoint sequences of a fixed length $l = 101$, where the first $l-1 = 100$ characters represent the input, and the final character represents the output. The loss is computed as the cross entropy loss between the expected character index and the output one.

\begin{figure*} [t!]
    \centering
    \includegraphics[width=\textwidth]{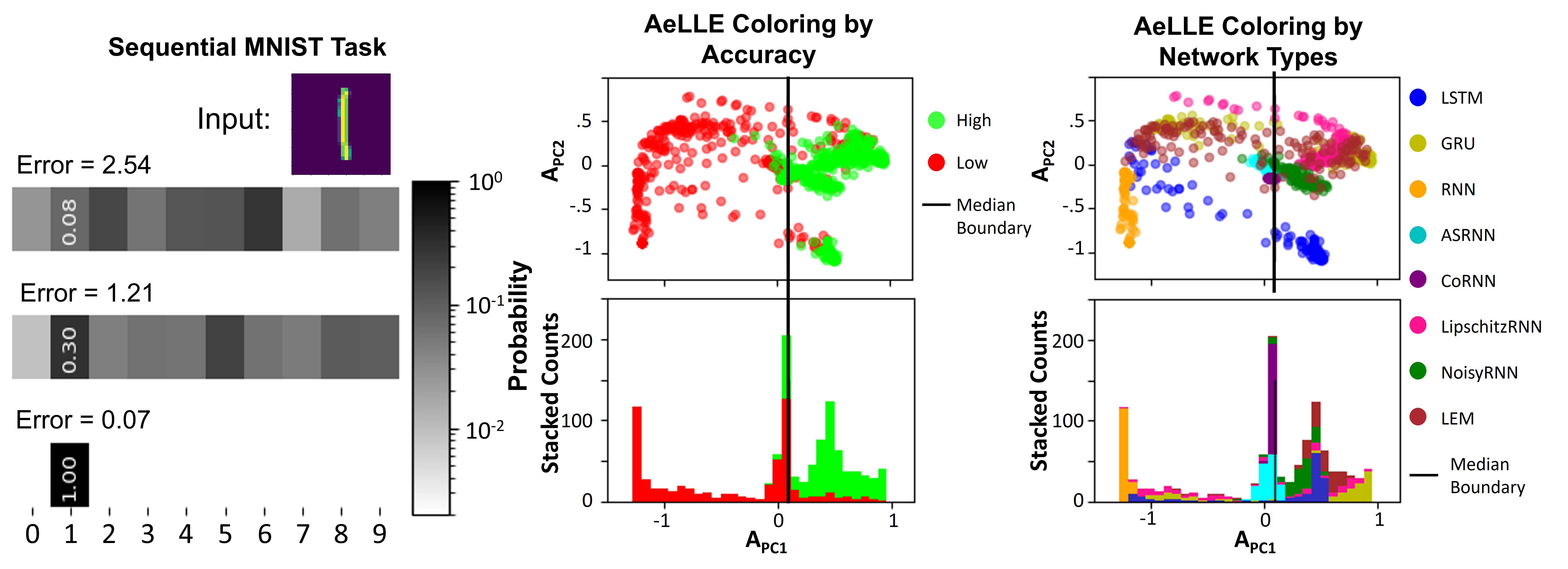}
        \caption{\textbf{Clustering across network architecture: AeLLE for networks trained on SMNIST Task.} \textit{Left:} In this task, the network predicts the digit shown in a given image. Higher losses correspond to lower confidence in the correct digit label. \textit{Center:} AeLLE representation shows that networks of the similar error are clustered together. When classifying these networks according to high accuracy (green) and low accuracy (red), the high-accuracy networks, regardless of network architecture, are consistently located further to the right (more positive PC1) than low-accuracy networks. \textit{Right:} Networks of the same type often clustered together in AeLLE representation, with some overlap between similar architectures (such as coRNN and ASRNN). For each network architecture (except vanilla RNN), there are variants with high and low accuracy separated across the first PC dimension.}
    \label{fig:SMNIST}
\end{figure*}

The hyperparameters of \textit{network size} (number of hidden states) and initialization of weight parameters appear to have most impact on the accuracy. We create 1200 variants of these parameters, varying the number of hidden units from 64 to 512 (by factors of 2) and sample initial weights from a symmetric uniform distribution with the parameter $p$ denoting the half-width of the uniform distribution from which the initial weights are sampled in the range of $[0.04,0.4]$. We split the variants into Autoencoder training set ($80\%$) and validation set ($20\%$).

Similar to the target learning task, we project the LE of the variant networks onto the first two PC dimensions of the latent space of the trained autoencoder and mark them according to accuracy. LSTM networks with loss below the median among these networks (loss $<$1.75) are considered as high-accuracy (\textit{green}), while those with loss above the median are considered low-accuracy (\textit{red}). 


We find that AeLLE in 2D space separates the spectra of the variants according to accuracy across network sizes. Performing principal component analysis of the AeLLE illustrates that the low- and high-accuracy networks are separated along the PC1 dimension, with higher-accuracy networks being further to the left in these space and lower-accuracy ones clustering to the right (Fig.~\ref{fig:CharRNN}). For comparison, we show the median value of the first principal component across all networks (black line), showing that the vast majority of high-accuracy networks are to the left of this line. In contrast, the distribution of the network sizes (Fig.~\ref{fig:CharRNN}-right) is more evenly distributed in this space. This demonstrates that this method is able to learn properties from the LE spectrum which correlate with performance across network sizes which are more informative than network size alone.

Comparing the separation in the AeLLE space with a classifier based on direct LE statistics, we find that using the median value of the mean Lyapunov Exponent or max Lyapunov Exponent as the decision boundary gives classifiers with F1 scores of 0.834 and 0.859, respectively, suggesting that both statistics are strongly indicative of performance on this task. The LE PC classifier with decision boundary defined by the median value of the first principal components of the raw Lyapunov exponents has a similar F1 score of 0.860. Meanwhile, the AeLLE classifier achieves an F1 score of 0.877, indicating an improvement on both of the direct statistics (see Supplementary Materials for more details). This shows that, while all metrics used are indicative of performance, the AeLLE method is able to achieve a slightly greater discrimination of network performance.

\subsection*{Sequential MNIST Classification with Different Network Types} \label{sec: SMNIST results}
A common benchmark for sequential models is the sequential MNIST task (SMNIST)~\cite{lecun1998gradient}. In this task, the input is a sequence of pixel greyscale values unrolled from an image of handwritten digits from $0-9$. The output is a prediction of the corresponding label (digit) written in the image. We follow the SMNIST task setup in~\cite{bai2018empirical}, where each image is treated as sequential data and each row is the input at one time, and the number of time steps is equal to the number of columns in the image. The loss corresponds to the cross entropy between the predicted and the expected one-hot encoding of the digit. 

We train a larger number of RNN variants on this task to demonstrate how the AeLLE properties translate across \textit{network architectures}. The architectures trained on this task were: LSTM, Gated Recurrent Unit (GRU), (vanilla) RNN, Anti-Symmetric RNN (ASRNN) \cite{Chang2019}, Coupled Oscillatory RNN (coRNN) \cite{rusch2021coupled}, Lipschitz RNN \cite{erichson2020lipschitz}, Noisy RNN \cite{lim2021noisy}, and Long-Expressive Memory Network (LEM) \cite{rusch2022long}. For each network type, we train 200 variants of hidden size 64. Every network was trained for 10 epochs and LE of post-trained networks are collected. This constitutes a set of $1600$ variants, where we use $70\%$ for Autoencoder training, $10\%$ for validation, and $20\%$ for testing.  For more details on the training of this Sequential MNIST task, see the Appendix A.3.
Similarly to previously described tests, we color code the variants according to accuracy. Networks with loss $<2.2\times 10^{-3}$ are considered as high-accuracy (\textit{green}) which includes $50\%$ of networks, while the rest of the networks with higher loss are considered as low-accuracy (\textit{red}).

\begin{figure*} [t!]
    \centering
    \includegraphics[width=0.9\textwidth]{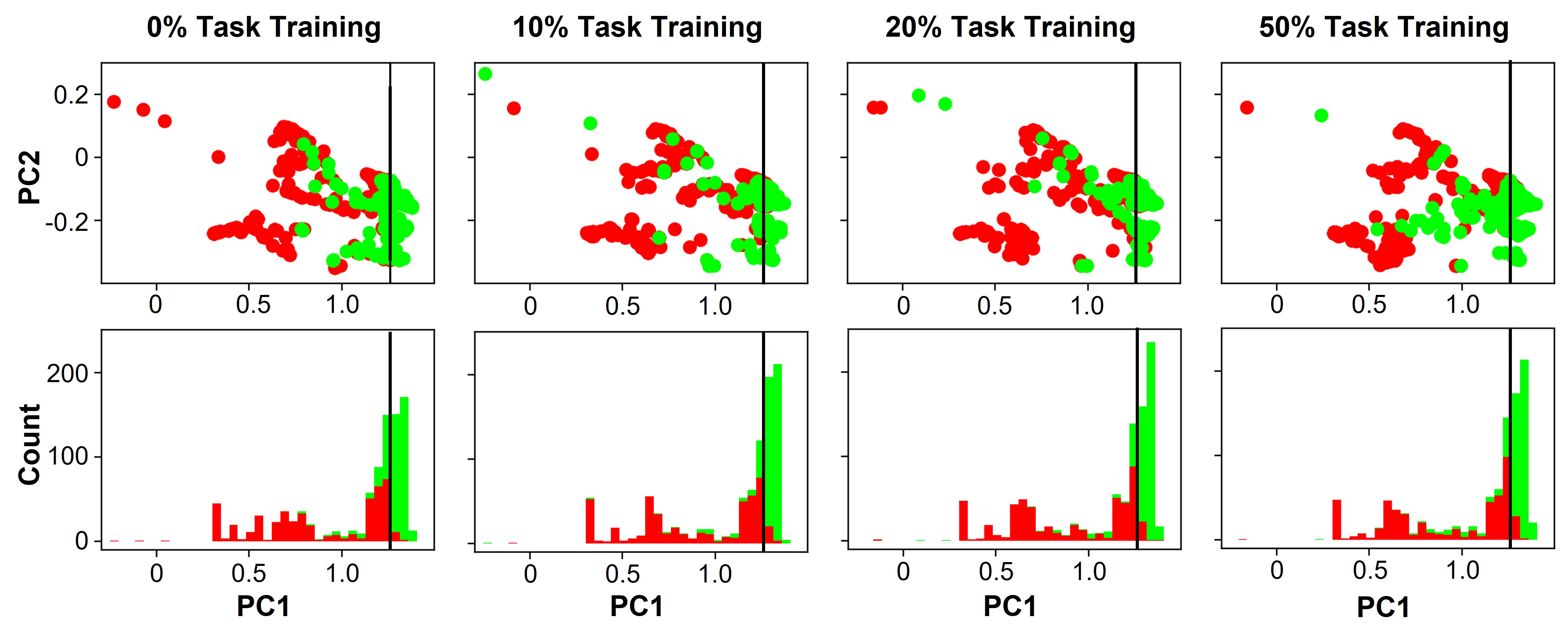}
    \caption{\textbf{Pre-Trained AeLLE during RNN training on SMNIST task.} Prediction of task accuracy of inputs in a test set by pre-trained AeLLE threshold classifier ($PC1 <-0.03$) of SMNIST variants is shown at ($0\%,10\%,20\%,50\%$) of training of networks on SMNIST. Throughout training, the high-accuracy networks (green) tend to cluster to the left of the Pre-trained AeLLE median classifier (black line). Furthermore, the distribution of networks in this space changes little over training, indicating that the dynamical properties which facilitate networks' successful learning of a task emerge early in training.}
    \label{fig:Extension}
\end{figure*}

As in previous tests, AeLLE analysis is able to unravel variants and their accuracy according to LE spectrum. With a median value of PC1 in AeLLE plane (black line in Fig.~\ref{fig:SMNIST}), the AeLLE plane could be divided into two clusters where low-accuracy are in the left part of the plane and high-accuracy are in the right part. The distribution of network architectures in this representation (see Fig.~\ref{fig:SMNIST}-right), we find clusters of mixed architectures with similar accuracy. Moreover, we find that GRU, Lipschitz RNN, LEM, and to a lesser extent Noisy RNN, all occupy a similar part of this space, with their best-performing variants generally being located to the top-right of the space shown, and then moving to the left as the perfomance of the variants deteriorates. While no vanilla RNN variants achieve high accuracy, even networks that have variants with high accuracy, such LSTM and LEM, have low-accuracy variants that are projected onto the same space as the cluster of vanilla RNNs. Meanwhile, ASRNN and coRNN, which are both constrained to have dynamics that preserve information, are projected very close to each other in this space into relative small clusters near the median boundary.   

The LE mean and LE max classifier on this dataset achieve F1 scores of 0.609 and 0.566, respectively, suggesting that both statistics are non-optimal predictors of performance on their own. The LE PC classifier has an F1 score of 0.608, representing a minor improvement in accuracy. However, the AeLLE classifier using the median value of the first principal component achieves F1 score of 0.859 (see Supplementary Materials for more details). This score is a significant improvement on the direct LE statistics and demonstrates that AeLLE is particularly advantageous for this task, suggesting that characteristics of the dynamics shared across architectures which determine network performance are non-trivial.


\subsection*{Pre-Trained AeLLE for Accuracy Prediction Across Training Epoch}
In the three tests described above, we find that the same general approach of AeLLE allows the selection of variants of hyperparameters of RNN associated with accuracy. LE spectrum is computed for fully trained models to set apart the sole role of hyperparameter variation. Namely, all variants in these benchmarks have been trained prior to computing LE spectrum. Over the course of training, connectivity weight parameters vary and as a result LE spectrum undergoes deformations. However, it appears that the general properties of LE spectrum such as the overall shape emerge early in training. 

From these findings and the success of AeLLE, a natural question arises: how early in training can AeLLE identify networks that will perform well upon completion of training? To investigate this question we use a pre-trained AeLLE classifier, i.e., trained on a subset of variants that were fully trained for the task. We then propose to test how such pre-trained fixed AeLLE represents variants that are only partially trained, e.g., underwent $0\%-50\%$ of training. This test is expected to show how robust are the inferred features within AeLLE correlating hyperparameters and LE spectrum subject to optimization of connectivity weights. Also it would provide insight into how long it is necessary to train the network to predict the accuracy of a hyperparameter variant.


\begin{table}[t!]
    \centering
    \caption{Precision, Recall, and F1 Score of pre-trained AeLLE classifier for RNN final accuracy evaluated at different stages of training. indicated by \textbf{bold numbers}. This classification is compared with a classification based on loss value at the corresponding epoch, indicated by [$\cdot$].}
    \begin{tabular}{l|l l|l|l}
        Training & \multicolumn{4}{c}{\textbf{AeLLE} vs. [Loss]} \\
        \hline
          \%& \multicolumn{2}{c|}{Recall} & Precision & F1 \\
         \hline
         0\% & 93.8\%& [--]& 77.0\%  [--] & 0.85  [--]\\
         10\% & \color{blue} \textbf{91.5\%} &\color{black} [16.9\%] & \textbf{86.6\%} \color{blue} [95.1\%] & \color{blue}  \textbf{0.89} \color{black}[0.35]\\
         20\% & \color{blue}\textbf{91.8\%} & \color{black}[45.8\%] & \textbf{86.5\%} \color{blue} [97.7\%] & \color{blue}  \textbf{0.90} \color{black}[0.66]\\
         50\% & \color{blue}\textbf{88.6\%} &\color{black}[77.7\%] & \textbf{83.7\%} \color{blue}  [97.3\%] & \color{blue} \textbf{0.86} \color{black}  [0.86] \\
    \end{tabular}

    \label{tab:PR_LEs}
\end{table}

We select the SMNIST task with LSTM, GRU, RNN, ASRNN, CoRNN, Lipschitz RNN, Noisy RNN, and LEM models with 64 hidden states size, the initialization parameter \textit{p} is the same as test three (200 variants for each model).
We then compute LE spectrum for the first five epochs of the training (out of all 10 epochs) for all variants. Note, LE spectrum before training are also computed. Therefore, $6000$ LE spectrum are considered. We then select $20\%$ variants into a training set, and they span over all epochs. We define such AeLLE as a Pre-Trained AeLLE and investigate its performance. 

\begin{figure*}[h!]
    \centering
    \includegraphics[width=0.9\textwidth]{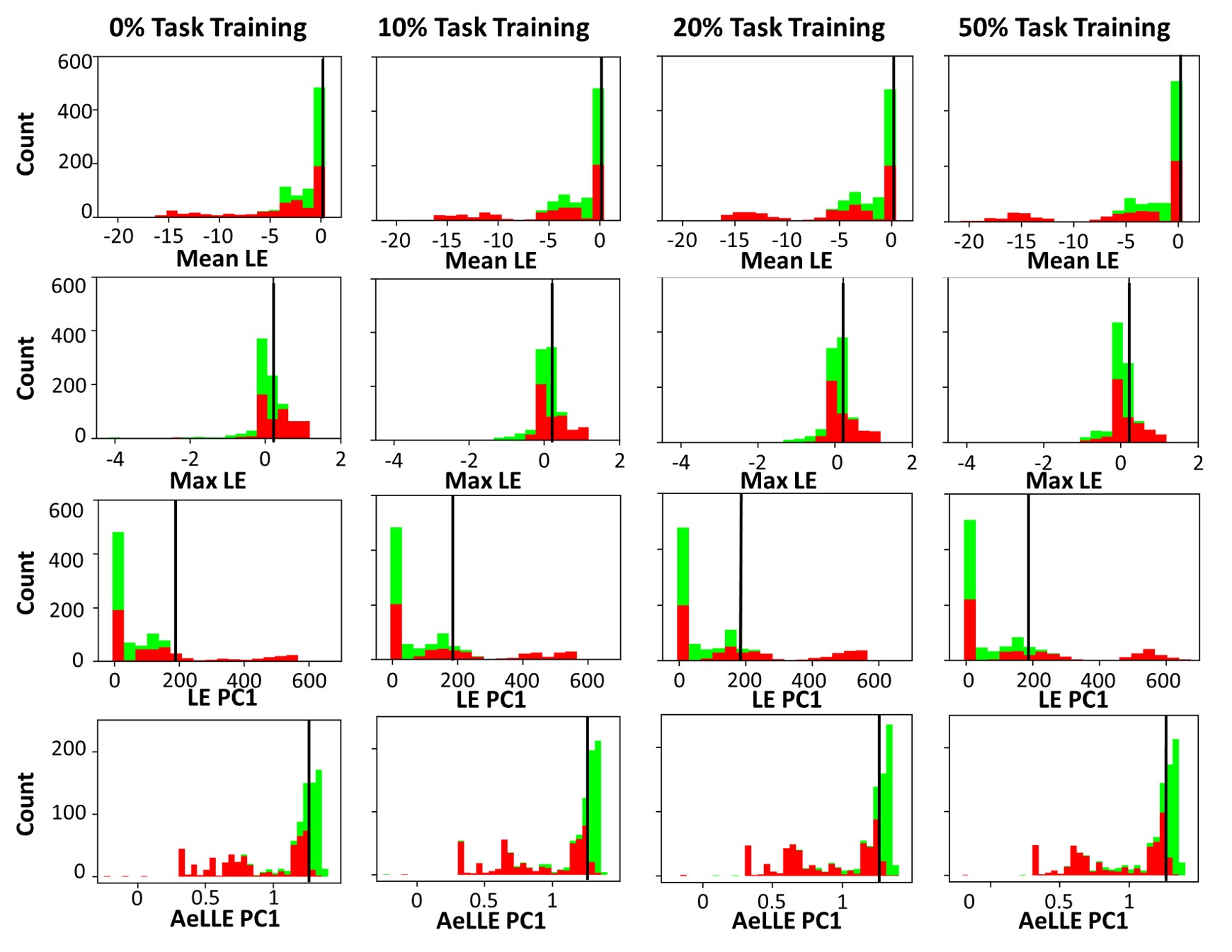}
    \caption{\textbf{Comparison of LE mean, LE max, and AeLLE classifiers and distributions throughout network training.} The distribution of the Mean LE (first row), Max LE (second row), and first principal component of the raw Lyapunov exponents (third row), and first principal component of the AeLLE (bottom) are shown from 0\% to 50\% training on the SMNIST task. The decision boundary value of each metric is shown as a vertical black bar, which is used as a classifier in Table \ref{tab:PR_LEstats}.}
    \label{fig:Prediction_LEs}
\end{figure*}
We select another $20\%$ data as the validation set ($1800$ variants), and apply the same loss threshold of $2.2\times10^{-3}$ as above. We use this validation set to define a simple threshold that classifies low- and high-accuracy variants according to AeLLE ( Fig.~\ref{fig:Extension} green shaded region of $PC1>-0.03$). 

We then apply the same Pre-Trained AeLLE and accuracy threshold to variants in the test set ($3600$ variants) at different epochs (formulated as \% of Training)  illustrated in Fig.~\ref{fig:Extension} with training progressing from left to right and Table~\ref{tab:PR_LEs}. We observe that projection of variants LE spectrum onto AeLLE (points in AeLLE 2D embedding) does not change significantly over the course of training. Specifically, before training, we find that the Recall, i.e., the number of the low-error networks which fall within the low-error threshold region, is $93.8\%$. The Precision, i.e., how many networks in the region are low-error, is $77.0\%$. Then after a single epoch, $10\%$ of training, the Recall becomes $91.5\%$ and the Precision improves to $83.6\%$, indicating more of the networks in the region are correctly identified as high-accuracy. The Recall and Precision does not change too much over training. This indicates that the LE spectrum captures properties of network dynamics which emerge with network initialization and remain throughout training.

In summary, we find that Pre-Trained AeLLE is an effective classifier that predicts the accuracy of the given RNN when fully trained, even before it has undergone 50\% training. To further quantify the effectiveness of AeLLE classifier prediction, we compare it with a direct feature of training, the loss value at each stage of training, see Table~\ref{tab:PR_LEs}. We find that variants with low loss early in training do correspond to variants that will be classified as low-error, indicated by almost perfect Precision rate of $99\%-100\%$. However, it appears that many variants of high accuracy do not converge quickly. Indeed, the Recall rate for a classifier based on loss values is $17\%-77\%$ for $10\%-50\%$ of training, while in contrast, AeLLE Recall rate is consistently above $88\%$ across all training epochs.

\begin{table}[h!]
    \centering

     
     \caption{Precision, Recall, and F1 Score of LE mean, LE Max, and AeLLE classifiers throughout training. The best of each score (Precision, Recall, F1) is indicated by \textbf{bold numbers}.}
    \begin{tabular}{l|l|l|l|l}
        Training & Classifier & Recall & Precision & F1 \\
         \hline
             & LE Mean & 99.1\% & 55.6\%  & 0.71 \\
         0\% & LE Max  & 65.3\% & \textbf{90.3\%}  & 0.72 \\
             & LE PCA  & \textbf{100\%} & 55.9\%  & 0.76 \\
             & AeLLE   & 93.8\% & 77.0\%  & \textbf{0.85} \\
         \hline
             & LE Mean & \textbf{97.7\%} & 58.3\%  & 0.73 \\
         10\%& LE Max  & 60.7\% & \textbf{92.6\%}  & 0.74 \\
             & LE PCA  & 96.8\%  & 60.0\%  & 0.74 \\
             & AeLLE   & 91.5\% & 86.6\%  & \textbf{0.89} \\
         \hline
             & LE Mean & \textbf{97.6\%} & 58.5\%  & 0.73 \\
         20\%& LE Max  & 59.5\% & \textbf{94.3\%}  & 0.73 \\
             & LE PCA  & 96.6\% & 60.3\%  & 0.74  \\
             & AeLLE   & 91.8\% & 86.5\%  & \textbf{0.90} \\
         \hline
             & LE Mean & \textbf{96.4\%} & 58.9\%  & 0.73 \\
         50\%& LE Max  & 58.8\% & \textbf{94.3\%}  & 0.72 \\
             & LE PCA  & 94.9\% & 60.4\%  & 0.74  \\
             & AeLLE   & 88.6\% & 83.7\%  & \textbf{0.86} \\
    \end{tabular}
    \label{tab:PR_LEstats}
\end{table}

For further comparison, we construct classifiers using the LE mean and LE max, and the first PC of the raw LEs, as training progresses. For each classifier, we select the value of the statistic that yields the best F1 score on the validation set as the decision boundary, and then test the accuracy of such a classifier in determining whether a network's performance at the end of training will be of high or low accuracy on the test set. The distribution of LE statistics and the first PC of the AeLLE classifier over training is shown in Figure \ref{fig:Prediction_LEs}. We see that the distribution of the Mean LE is heavily concentrated near its maximum value of 0 and does not change significantly over training. Max LE distribution exhibits more changes, but there is a similar number of high- and low-accuracy networks to either side of the boundary throughout training. The first PC of AeLLE has the greatest concentration of high-accuracy networks to one side of the decision boundary, suggesting that this is a more useful classifier. In Table \ref{tab:PR_LEstats}, we present the recall, precision, and F1 score of each of these classifiers. This comparison shows that the AeLLE classifier has significantly higher F1 scores than the other classifiers across all epochs.

\subsubsection*{AeLLE Contribution to Classification}
To delineate the necessity of AeLLE representation, we extend the classifier from a simple classifier (PC1), which we used in earlier sections, to a linear regression classifier. We then perform a comparative study to test whether AeLLE representation is responsible for the effective classification that we observe or more advanced classifier applied directly to LE could achieve a similar outcome. To test this, we use a linear regression classifier in conjunction with AeLLE and compare the classification accuracy with a scenario of using the classifier directly with LE (both full spectrum (Raw) and low dimensions (PCA)). Table~\ref{tab:linear_regression} shows the results of the study, in terms of F1 score, and demonstrates that AeLLE is a required step in classification.  In particular, in the two scenarios where AeLLE representation is used, classification in conjunction with PC1 or in conjunction with linear regression achieves similar F1 scores (last two rows in Table~\ref{tab:linear_regression}). These scores are significantly higher than the application of linear regression directly to LE values (row 1) or to LE values projected to lower dimensional space with PCA (rows 2, 3). We find that for all tested training durations (from 0\%-50\%), classifiers applied to AeLLE achieve an F1 score on average higher by \textit{0.12} than the score computed directly on LE or its PCA embedding. In addition, the study suggests that an extension of the classifier, such as extending the classifier from PC1 to linear regression, does not replace AeLLE but rather could provide additional means in conjunction with AeLLE to enhance classification.

\begin{table*}[t]
    \centering
    \caption{Comparison of F1 score of PC1 and Linear Regression classifiers in conjunction with LE-related representations (Raw - row 1, PCA - rows 2,3 , AeLLE - rows 4,5) throughout training duration (from 0\% training to 50\% ). The testing set consists of all samples from 0\% training to 50\% training. The highest score for each training duration is marked in bold and classifiers in conjunction with AeLLE are marked in blue.}
    \small
    \begin{tabular}{l|c|c|c|c|c|c|c}
    Repr Space and Clf & 0\% & 10\% & 20\% & 30\% & 40\% & 50\% & Testing set\\
    \hline
    LE (Raw) - Lin. Reg.~ & 0.62 & 0.61 & 0.61 & 0.62 & 0.64 & 0.61 &0.62\\
    \hline
    LE (PCA) - PC1 & 0.72 & 0.74 & 0.74 & 0.75 & 0.75 & 0.74 & 0.74\\
    \hline
    LE (PCA) - Lin. Reg. & 0.73 & 0.77 & 0.78 & 0.78 & 0.77 & 0.75 & 0.76\\
    \hline \hline
    AeLLE (PCA) - PC1~ & \color{blue} 0.85 &  \color{blue} \textbf{0.89} & \color{blue} \textbf{0.89} & \color{blue} \textbf{0.90} & \color{blue} \textbf{0.89} & \color{blue} 0.86 & \color{blue} 
    \textbf{0.88}\\
    \hline
    AeLLE (PCA) - Lin. Reg. & \color{blue} \textbf{0.86} &  \color{blue}\textbf{0.89} &  \color{blue}\textbf{0.89} &  \color{blue}\textbf{0.90} &  \color{blue}\textbf{0.89} &  \color{blue}\textbf{0.87} &  \color{blue}\textbf{0.88} 
    \end{tabular}
    \label{tab:linear_regression}
\end{table*}

\begin{table}[b]
    \centering
    \caption{F1 Score of pretrained AeLLE classifier on networks at various levels of training using median from increasing numbers of principal components (PCs). The total dimension of the AeLLE space is 32.}
    \begin{tabular}[c]{c|c|c|c|c}
    & \multicolumn{4}{c}{\#PCs} \\ \hline
\textbf{\bf Training \%} & 1 & 2 & 4 & 10\\ \hline
0\%  & 0.81 & 0.81 & 0.84 & 0.89 \\\hline
50\% & 0.84 & 0.84 & 0.87 & 0.91 \\\hline
    \end{tabular}

    \label{tab:alternative_classifier}
\end{table}

\subsubsection*{Higher Dimensions of AeLLE}
AeLLE space is not restricted to two dimensions. In general, the more dimensions are used, the representation is expected to become more accurate. To test this hypothesis we extend the classification to higher dimensions. Specifically, we use the first $d$ PCs where the median of each PC divides the space into $2^d$ subspaces. In each subspace, we check the number of optimal and non-optimal network samples. Then, we take the union of all optimal subspaces (those which contain more optimal networks than sub-optimal networks) to be the overall optimal region for the classifier. We then test the classifier on the test dataset for each epoch. Our results are reported in Table~\ref{tab:alternative_classifier}, where we show the F1 score for $0\%$ and $50\%$ training with $d$ being the number of PCs is set to $d=1,2,4,10$.

The results show that as additional PCs are included, F1 score increases at initial phase before training and during training, hence the full AeLLE space includes in higher dimensions additional features of LE and the corresponding networks. The results also show that the first-order PC classifier is able to capture a major portion of high- and low-accuracy networks.

\begin{figure*}[t!]
    \centering
    \includegraphics[width=1.0\textwidth]{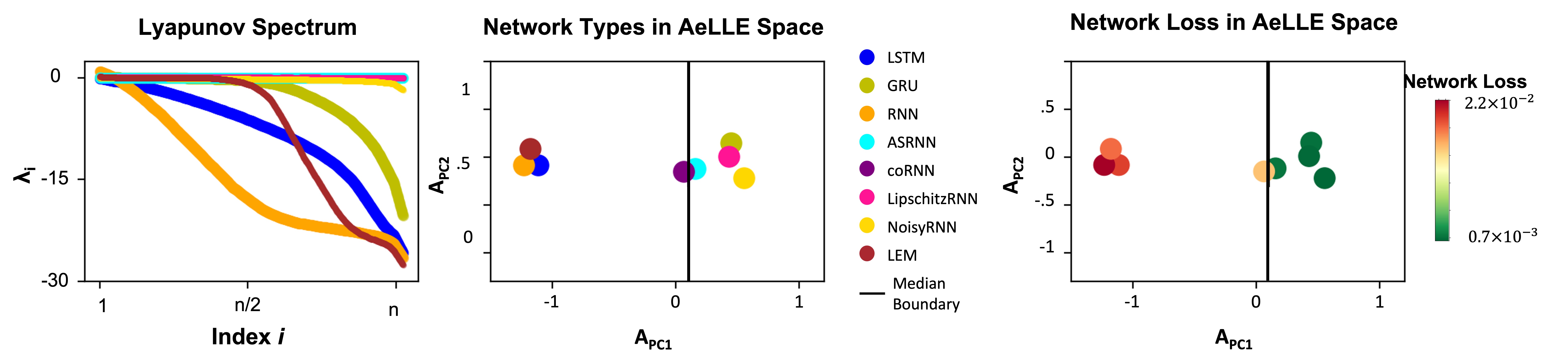}
    \caption{\textbf{Comparison of Lyapunov Spectrum curves, AeLLE, and loss for different network architectures on SMNIST task.} Lyapunov Spectrum curves (left) of five different architectures show that the ASRNN, coRNN, Lipschitz RNN, and Noisy RNN have very similar spectra with many exponents close to zero. Meanwhile, the spectra of LSTM, GRU, RNN, and LEM dip well below zero, but at different rates. These networks in AeLLE space (center), are thus grouped such that RNN, LSTM, and LEM are closer to each other than to other network types, but that the GRU is grouped with the networks with spectra close to zero. Such mapping appears to be correlated with accuracy (right; colors indicate loss from low-green to high-red). Despite the similar spectra of coRNN, ASRNN, Lipschitz RNN, and Noisy RNN the loss of coRNN is larger than that of the others by an order of magnitude, causing it to be separated from some of the other networks in this space. While the LE curves of LSTM, RNN and LEM are distinct, their loss values are much higher (red) and they are clustered in the same vicinity farther from more optimal networks.}
    \label{fig:LS_samples}
\end{figure*}

\subsection*{LE Features Visualization in AeLLE Space}

Lyapunov Spectrum for different architectures can have highly variable maximum values, slopes, variances, means, and more. Whereas certain properties of these spectra would intuitively correlate with performance, the relation between each individual feature and accuracy is unclear. This is evident from comparison of Lyapunov Spectra of LSTM, RNN, GRU, and LEM to coRNN, ASRNN, Lipschitz RNN, and Noisy RNN trained on SMNIST in Figure \ref{fig:LS_samples}(left). Whereas coRNN, ASRNN, Lipschitz RNN, and Noisy RNN all have exponents close to zero for all indices, the spectra of LSTM, GRU, RNN, and LEM all dip well below zero, with RNN decreasing much faster than the others. Meanwhile, GRU maintains a far greater number of exponents close to zero, similar to coRNN, for about half of the indices. It would then be natural to assume the performance of all networks with all LEs close to zero would be most similar and that LSTM, GRU, and GRU would all be most similar to RNN in performance.

The representation in AeLLE space (Figure \ref{fig:LS_samples} center) shows two clusters, with one tight cluster representing RNN, LSTM, and LEM and another looser cluster with all other sample networks. Within this looser cluster, we find both the coRNN network, which has a very similar spectrum to ASRNN and Lipschitz RNN, and the GRU, which is visually much closer to LSTM and LEM. However, we find that, within this cluster, GRU is located closest to Lipschitz RNN and Noisy RNN, all of which are found to the right of the median boundary along with ASRNN. Meanwhile, coRNN is located just to the left of the boundary, in the direction of the RNN, LSTM, and LEM. While visualy inspection of the spectra does not immediately indicate the reason for this ordering, it becomes more clear when we observe the loss of the networks (Figure \ref{fig:LS_samples} right). Since ASRNN, Lipschitz RNN, Noisy RNN, and GRU obtain the optimal accuracy (indicated by green color), they are mapped to the right of coRNN and the other networks point in AeLLE plane. Furthermore, while LSTM, RNN, and LEM are different networks in architectures and dynamics, these appear to be mapped to the same cluster in AeLLE. The reason for such non-intuitive mapping could be explained by accuracy again, since on this task, RNN LSTM, and LEM all exhibit low accuracy (indicated by red color).

Such experiments indicate that easily observable LE features such as number of exponents near zero or overall spectrum shape do not vary uniformly with performance. Instead,  more complex, non-linear combination of LE features extracted by AeLLE would be required to determine this relation.

\section*{Discussion}

LE methodology is an effective toolset to study nonlinear dynamical systems since LE measure the divergence of nearby trajectories, and thus indicate the degree of stability and chaos in that system. Indeed, LE has been applied to various dynamical systems and applications, and there exist theoretical underpinning for characterization of these systems by LE spectrum. However, it is unclear how to relate LE spectrum to system characterization. Our results demonstrate that the information that LE contain regarding the dynamics of a network can be related to network accuracy on various tasks through an autoencoder embedding, called AeLLE. In particular, we show that AeLLE representation encodes information about the dynamics of recurrent networks (represented by their Lyapunov Exponents) along with the performance. We demonstrated that this relation to performance can be learned across choice of  weight initialization, network size, network architecture, and even training epoch. Effectively, AeLLE representation discovers the implicit parameters of the network.

Such a representation is expected to be invaluable in uniting research which looks to assess and predict model quality on particular tasks and those which emphasize and constrain model dynamics to encourage particular solutions. Our approach allows mapping of dynamics of a network to accuracy through the latent space representation of LE autoencoder. This mapping appears to capture multiple characteristics of the networks, with some are direct such as network type, accuracy, number of units, and some are implicit. All these appear to be be contained in AeLLE representation and effectively provide salient features/parameters for the networks that are being considered.

Specifically, the significance of our results is that they show that AeLLE representation is able to distinguish networks according to accuracy across choice of network architectures with greater accuracy than using simple spectrum statistics alone.  Such findings suggest that the features of the dynamics which correspond to optimal performance on given tasks are consistent across network architectures, whether they are RNN, gated architectures, or dynamics-constrained architectures, even if they are not immediately apparent upon visual inspection or through first-order statistics of the spectrum. AeLLE is able to capture these.

While the accuracy of such a classifier is enhanced over those using direct statistics, this comes at the cost of training time of the autoencoder and interpretability of extracted features. Analysis of the individual components of this autoencoder methodology (\eqref{eq:AE}) could provide more insight into the interpretation of these AeLLE features. Namely, the representations of the Lyapunov exponents which AeLLE produces ($\phi (Z)$) could be directly analyzed statistically or otherwise. The contribution of each dimension of the latent space of the autoencoder to the predicted loss value could be extracted from the linear prediction layer ($\xi$). Furthermore, the corresponding LE spectrum features for these latent dimensions could be analyzed using the decoder ($\psi$). Additional experiments exploring the AeLLE representation beyond those outlined here could also be carried out to provide further insights. These studies are left for future work to be applied with optimization and analysis of particular machine learning tasks, based on generic methodology presented in this work.


Furthermore, while the application of AeLLE for search of optimal networks given a task is outside of the scope of this work, exploration into AeLLE representation to find predicted optimal dynamics for a task would be a natural extension of the results that are reported here. Furthermore, extension of AeLLE methodology can be used to search and unravel novel architectures with desired dynamics defined by a particular LE spectrum. AeLLE approach can be also adopted to analyze other complex dynamical systems. For example, long-term forecasting of temporal signals from dynamical systems is a challenging problem that has been addressed with a similar data-driven approach using autoencoders and spectral methods along with linearization or physics-informed constraints ~\cite{Lusch2018, lange2021fourier, morton2018deep, erichson_autoencoders, azencot2020forecasting}. Application of AeLLE could unify such approaches for dynamical systems representing various physical systems. The key building blocks in AeLLE that would need to be established for each of these extensions is efficient computation of Lyapunov exponents, and sufficient sampling of data to train the Autoencoder to form an informative Latent representation, and stable back-propagation of gradients across many iterations of QR decomposition in the LE calculation.

\section*{Code and Data Availability}
The datasets analysed during the current study as well as the code used to generated results are available in the LyapunovAutoEncode repository, https://github.com/shlizee/LyapunovAutoEncode.
\bigskip


\bibliography{ref}

\end{document}